\documentclass[preprint,12pt]{elsarticle}

\usepackage{amssymb}
\usepackage{amsmath}

\usepackage{multirow}
\usepackage{tabularx}
\usepackage{url}

\journal{...}

\begin{document}

\begin{frontmatter}



\title{Advancing ALS Applications with Large-Scale Pre-training: Dataset Development and Downstream Assessment}



\author{Haoyi Xiu}
\author{Xin Liu}
\author{Taehoon Kim}
\author{Kyoung-Sook Kim}
\affiliation{organization={National Institute of Advanced Industrial Science and Technology (AIST)\\ Artificial Intelligence Research Center},
            city={Tokyo},
            country={Japan}}

\begin{abstract}
The pre-training and fine-tuning paradigm has revolutionized satellite remote sensing applications. However, this approach remains largely underexplored for airborne laser scanning (ALS), an important technology for applications such as forest management and urban planning. In this study, we address this gap by constructing a large-scale ALS point cloud dataset and evaluating its impact on  downstream applications. Our dataset comprises ALS point clouds collected across the contiguous United States, provided by the United States Geological Survey's 3D Elevation Program. To ensure efficient data collection while capturing diverse land cover and terrain types, we introduce a geospatial sampling method that selects point cloud tiles based on land cover maps and digital elevation models. As a baseline self-supervised learning model, we adopt BEV-MAE, a state-of-the-art masked autoencoder for 3D outdoor point clouds, and pre-train it on the constructed dataset. The pre-trained models are subsequently fine-tuned for downstream tasks, including tree species classification, terrain scene recognition, and point cloud semantic segmentation. Our results show that the pre-trained models significantly outperform their scratch counterparts across all downstream tasks, demonstrating the transferability of the representations learned from the proposed dataset. Furthermore, we observe that scaling the dataset using our geospatial sampling method consistently enhances performance, whereas pre-training on datasets constructed with random sampling fails to achieve similar improvements. These findings highlight the utility of the constructed dataset and the effectiveness of our sampling strategy in the pre-training and fine-tuning paradigm. The source code and pre-trained models will be made publicly available at \url{https://github.com/martianxiu/ALS_pretraining}.
\end{abstract}




\end{frontmatter}



\section{Introduction}\label{sec: intro}
Airborne Laser Scanning (ALS) is an important remote sensing technology that captures high-resolution, three-dimensional spatial data by emitting laser pulses from an airborne platform and analyzing the reflected signals. This process generates dense light detection and ranging (LiDAR) point clouds, which accurately represent the Earth’s surface including both natural and built environments. A significant advantage of ALS is its ability to penetrate vegetation and provide precise measurements, making it particularly valuable for applications such as terrain mapping~\cite{qin2021opengf}, forest management~\cite{pureforest}, urban planning~\cite{dales}, and disaster management~\cite{xiu2023ds}.

Large-scale pre-training and fine-tuning paradigms have been transformative across various artificial intelligence (AI) fields~\cite{brown2020language,radford2021learning}. These paradigms involve extensive pre-training on diverse datasets, enabling models to adapt effectively to a wide range of downstream tasks through fine-tuning. Commonly referred to as foundation models~\cite{bommasani2021opportunities}, they leverage large-scale self-supervised/unsupervised training to learn generalizable representations.
Satellite remote sensing has also greatly benefited from this trend. By pre-training on large-scale unlabeled datasets, such as Sentinel-2, remote sensing foundation models~\cite{skysense,GFM,spectralgpt} achieve state-of-the-art performance on a variety of downstream tasks, including scene classification, land cover mapping, and multi-temporal cloud imputation.

However, large-scale pre-training and fine-tuning paradigms have yet to demonstrate their full impact on ALS applications. Although several large-scale datasets, such as OpenGF~\cite{qin2021opengf} and PureForest~\cite{pureforest}, exist, they lack the scale and land cover diversity necessary for training versatile models. Furthermore, while numerous freely available ALS LiDAR data sources, such as the United States Geological Survey 3D Elevation Program (USGS 3DEP)~\cite{stoker2022accuracy} and the Actueel Hoogtebestand Nederland (AHN)~\cite{ahn_website}, provide extensive resources, there is currently no efficient method to extract data from these sources. Leveraging all available data is computationally prohibitive and often redundant. These limitations collectively hinder progress in adopting the pre-training and fine-tuning paradigm for ALS applications. 

To address the aforementioned limitations, this study focuses on developing a large-scale dataset for pre-training ALS models and evaluating its effectiveness on downstream tasks. First, we propose a geospatial sampling method to extract point cloud tiles from the extensive resources provided by 3DEP. Our sampling method leverages land cover maps from the National Land Cover Database (NLCD)~\cite{wickham2023thematic} and the USGS seamless digital elevation models (DEM), aiming to maximize land cover and terrain diversity—factors critical for ALS applications.
Second, to assess the utility of the constructed dataset, we perform pre-training and fine-tuning on two downstream tasks. We adopt BEV-MAE~\cite{bevmae}, a self-supervised learning (SSL) model designed for outdoor 3D point clouds, as our baseline due to its state-of-the-art performance and suitability for ALS data (see Section~\ref{sec: model} for details). The model is fine-tuned on several  downstream applications including tree species classification, terrain scene recognition, and point cloud segmentation.
Additionally, we demonstrate the effectiveness of the proposed sampling method by scaling up the dataset and monitoring relative performance improvements, comparing it against alternative datasets and random sampling methods.

In summary, our contributions are as follows:
\begin{itemize}
    \item We construct a large-scale dataset for SSL on ALS point clouds and introduce a geospatial sampling method that leverages land cover maps and digital elevation models for efficient and diverse data collection.
    \item We pre-train and fine-tune models on the constructed dataset to evaluate the utility of the developed datasets, sampling methods, and pre-trained models. Additionally, to assess performance in recognizing different terrain scenes, we create a terrain scene recognition dataset based on an existing dataset designed for ground filtering.
    \item We release the code, dataset, and pre-trained models to the community, with the goal of advancing the pre-training and fine-tuning paradigm in ALS research.
\end{itemize}
 
The rest of the paper is organized as follows: In Section~\ref{sec: related work}, we review and summarize related work. Section~\ref{sec: data} describes the development of our dataset, including the detailed design of the sampling method and key statistics of the constructed dataset. In Section~\ref{sec: model}, we provide details about the model used in this study. Subsequently, Section~\ref{sec: experiments} and Section~\ref{sec: results} cover the experimental design and results, respectively. Finally, we conclude the study in Section~\ref{sec: conclusion}.

\section{Related Work}\label{sec: related work}
\subsection{Remote sensing foundation models}

Recently, AI has experienced a major paradigm shift from the supervised learning paradigm to the pre-train--fine-tune paradigm, where a large model is pre-trained with self-supervision on large-scale data and subsequently fine-tuned on downstream or target tasks~\cite{bommasani2021opportunities}. These pre-trained models, known as foundation models, are designed to be broadly applicable across a wide range of tasks with minimal adaptation. Prominent examples include GPT-3~\cite{brown2020language}, CLIP~\cite{radford2021learning}, and LLaVA~\cite{liu2024visual}.

The remote sensing community has quickly embraced this trend, with many researchers exploring the potential of foundation models in this domain. Recently, numerous vision-based foundation models for remote sensing have emerged. Early works adopted contrastive learning approaches to construct foundation models without the need for annotations~\cite{ayush2021geography,seco, caco}. These methods often extend existing approaches from computer vision while incorporating characteristics unique to satellite imagery. For example, Ayush et al.~\cite{ayush2021geography} adapted MoCo-v2~\cite{mocov2} for remote sensing data by reformulating the pretext task to utilize geolocation and temporal image pairs. SeCo~\cite{seco} designed self-supervision tasks that leverage the seasonal and positional invariances in remote sensing data, acquiring representations invariant to seasonal and synthetic augmentations. Similarly, CaCo~\cite{caco} introduced a novel objective function that contrasts long- and short-term changes within the same geographical regions. More recently, SkySense~\cite{skysense} was introduced as a billion-scale foundation model pre-trained on multitemporal optical and SAR images. It adopts multi-granularity contrastive learning to capture representations across different modalities and spatial granularities, achieving state-of-the-art (SoTA) performance on seven downstream tasks.

In addition to contrastive learning, numerous approaches based on MAE~\cite{mae} have also been proposed. For instance, SatMAE~\cite{satmae} incorporates temporal and spectral position embeddings to effectively utilize temporal and multispectral information. RingMo~\cite{ringmo} introduces a novel masking strategy to better preserve dense and small objects during masking. GFM~\cite{GFM} leverages the strong representations learned from ImageNet-22k~\cite{imagenet} and enhances remote sensing image representation through continual pre-training. To better use spectral information in satellite imagery, SpectralGPT~\cite{spectralgpt} introduces a 3D masking strategy and spectral-to-spectral reconstruction. Meanwhile, msGFM~\cite{msGFM} incorporates cross-sensor pre-training using four different modalities to learn unified multi-sensor representations. This approach outperforms single-sensor foundation models across four downstream datasets.

Research into vision-language models (VLMs) is also highly active, as these models enable zero-shot applications. For instance, RemoteCLIP~\cite{remoteclip} and SkyCLIP~\cite{skyscript} adapt CLIP for remote sensing datasets, outperforming standard CLIP baselines. GRAFT~\cite{graft} introduces a pre-training framework that uses ground images as intermediaries to connect text with satellite imagery, enabling pre-training without textual annotations. GeoChat~\cite{kuckreja2024geochat} fine-tuned LLaVA-1.5~\cite{liu2024improved} on a proposed instruction-following dataset, demonstrating promising zero-shot performance across a wide range of tasks, including image and region captioning, visual question answering, and scene classification.

Despite the significant advancements in developing foundation models for satellite imagery, to the best of our knowledge, no prior work has investigated foundation models specifically for ALS data—an area we aim to explore in this study.

\subsection{Datasets for 3D geospatial applications}
With the rapid advancements in 3D acquisition technologies, the availability of outdoor point cloud datasets has grown significantly, driving progress in 3D geospatial data analysis through deep learning techniques. Existing datasets can be broadly categorized based on their data collection methods. Photogrammetric 3D datasets, such as Campus3D~\cite{li2020campus3d}, SensatUrban~\cite{hu2022sensaturban}, HRHD-HK~\cite{li2023hrhd}, and STPLS3D~\cite{chen2022stpls3d}, are generated using photogrammetry techniques but lack ground points beneath dense vegetation canopies due to the limitations of passive image capture, making them unsuitable for ALS applications requiring dense vegetation analysis. Terrestrial and mobile laser scanning (TLS/MLS) datasets, including Semantic3D~\cite{hackel2017semantic3d}, Paris-Lille-3D~\cite{roynard2018paris}, SemanticKITTI~\cite{behley2019semantickitti}, and Toronto-3D~\cite{tan2020toronto}, are collected at street level and focus on roadway scene understanding. While they provide high point density and large data volumes, their limited geographic coverage, as well as restricted diversity in land cover and terrain, making them inadequate for broader ALS applications. ALS datasets, such as ISPRS Vaihingen 3D~\cite{rottensteiner2014results}, DublinCity~\cite{zolanvari2019dublincity}, LASDU~\cite{ye2020lasdu}, DALES~\cite{dales}, and OpenGF~\cite{qin2021opengf}, are collected using airborne LiDAR sensors and primarily target urban classification and environmental perception by identifying common urban objects like ground, grass, fences, cars, and facades. However, their limited scale and coverage restrict their utility for training versatile models across diverse ALS tasks. 

Unlike these datasets, which are designed with specific applications in mind, this study aims to construct a large-scale dataset encompassing a wide range of terrains and land cover types to support the pre-training of generalizable 3D models tailored for ALS applications.

\subsection{SSL methods for 3D Point Clouds}
\subsubsection{SSL methods for general 3D point clouds}
SSL enables neural networks to learn from unlabeled data, making it ideal for 3D point clouds where annotations are costly. We mainly focuses on masked autoencoding based SSL methods as it is most relevant to this study. Masked autoencoding methods learn meaningful representations by reconstructing randomly masked input portions, such as image patches or text tokens, capturing structural and contextual information.

Early works focus on generalizing BERT~\cite{devlin2018bert}’s masked language modeling to point clouds~\cite{yu2022point,fu2024pos,fu2023boosting}. A representative example is Point-BERT~\cite{yu2022point}, which trains a transformer encoder to predict masked dVAE~\cite{rolfe2017discretevariationalautoencoders}-generated tokens. Following the proposal of MAE, this idea was extended to point clouds. For example, Point-MAE~\cite{pang2022masked} applies the concept by treating local point neighborhoods as patches for reconstruction. MaskPoint~\cite{liu2022masked} introduces a masked discrimination task, replacing reconstruction with real/noise discrimination to improve robustness against sampling variance. Point-M2AE~\cite{zhang2022point} incorporates a hierarchical masking strategy for multi-scale pre-training, while PointGPT~\cite{chen2024pointgpt} adopts GPT-style generative pre-training with a decoder tasked to generate point patches. MaskFeat3D~\cite{yan20233d} instead reconstructs surface properties, like normals, to learn higher-level features. 

Some methods explore multi-modal approaches to enhance representation quality. I2P-MAE~\cite{zhang2023learning} incorporates 2D-guided masking and reconstruction using knowledge from pre-trained 2D models. Joint-MAE~\cite{guo2023joint} performs joint masking and reconstruction of 2D and 3D data with shared encoders and decoders. RECON~\cite{qi2023contrast} combines masked autoencoding and contrastive learning, leveraging their respective strengths while handling inputs from points, images, and text.

Extensive work has focused on autonomous driving. Unlike indoor or synthetic point clouds, outdoor point clouds are sparse and have varying density. Traditional masked point modeling strategies often create overlapping patches, discarding important points. To address this, Voxel-MAE~\cite{hess2023masked} uses a voxel-based masking strategy, predicting point coordinates, point counts per voxel, and voxel occupancy to better capture outdoor data distributions. Geo-MAE~\cite{tian2023geomae} improves further by predicting centroids, surface normals, and curvatures. GD-MAE~\cite{yang2023gd} introduces a generative decoder, eliminating the need for complex decoders or masking strategies. Recently, BEV-MAE~\cite{bevmae} explicitly focuses on learning BEV representations, achieving superior and efficient performance.

In this work, we focus on the impact of pre-training on downstream tasks for ALS data rather than designing a new network. We adopt BEV-MAE as it suits ALS data well, with details provided in Section~\ref{sec: model}.

\subsubsection{SSL methods for ALS 3D point clouds}
SSL has recently been applied to ALS. \cite{caros2023self} uses Barlow Twins~\cite{zbontar2021barlow} to improve semantic segmentation, especially for under-represented categories. \cite{de2023deep} proposes a deep clustering and contrastive learning approach for unsupervised change detection, outperforming traditional methods. \cite{yang2024self} pre-trains customized transformers under the MAE framework for 3D roof reconstruction, surpassing general MAE-based methods like Point-MAE and Point-M2AE. HAVANA~\cite{zhang2024havana} enhances contrastive learning by improving negative sample quality through AbsPAN.

While effective in their respective applications, none have developed large-scale datasets or conducted large-scale pre-training on ALS point clouds for building general-purpose models, which is our primary goal.

\section{Data for pre-training}\label{sec: data}
\begin{figure}[htb]
    \centering
    \includegraphics[width=\textwidth]{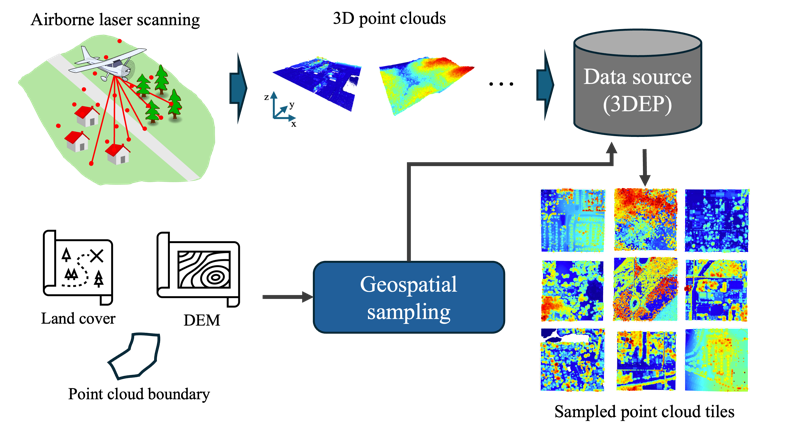} 
    \caption{Overview of the dataset development procedure: Land cover data, DEM, and point cloud boundaries are used to selectively download point cloud tiles from a remote server provided by 3DEP. The point clouds are visualized with elevation-based coloring, where cooler colors represent lower elevations and warmer colors indicate higher elevations.} 
    \label{fig:dataset_flow}
\end{figure}

In this section, we present the data source, describe the methodology used to develop the dataset, and provide key statistics of the resulting dataset to offer valuable insights into its characteristics. The overall procedure is shown in Figure~\ref{fig:dataset_flow}.

\subsection{Data source}
Since ALS is often used for extracting objects on the Earth’s surface or terrain information, the pre-training dataset should encompass diverse types of land cover and terrain to ensure that the trained models achieve strong generalizability.

In this work, we use LiDAR data from the USGS 3DEP~\cite{usgs_3dep} as the primary data source for building a large-scale dataset for pre-training. 3DEP is a collaborative program designed to accelerate the collection of three-dimensional (3D) elevation data across the United States to meet a wide variety of needs~\cite{stoker2022accuracy}. High-quality LiDAR data are collected for the conterminous United States (CONUS), Hawaii, and U.S. territories, while interferometric synthetic aperture radar (IfSAR) data are collected for Alaska. The program has established specifications for collecting 3D elevation data and developed data management and delivery systems to ensure public access to these datasets in open formats~\cite{stoker2022accuracy}. The data are readily accessible through tools such as LiDAR Explorer~\cite{usgs_lidar_explorer} and The National Map~\cite{usgs_national_map_downloader}. These datasets have been widely used to support U.S. local economies~\cite{cretini20233d, rachol20243d, fredericks20243d} and advance scientific research~\cite{chirico2020evaluating, oh2022high, scott2022statewide}.

\begin{figure}[htb]
    \centering
    \includegraphics[width=\textwidth]{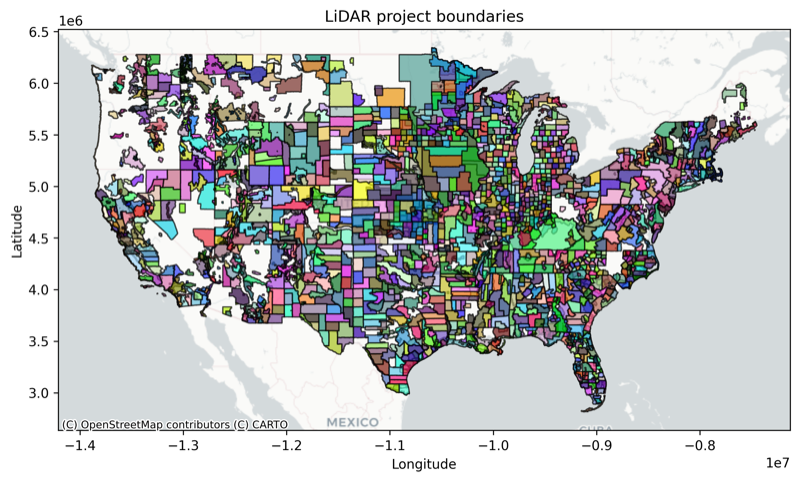} 
    \caption{LiDAR point cloud boundaries used in this study are shown with randomly assigned colors for the boundary polygons. The boundary data were downloaded from~\cite{usgs_3dep_lidar_aws} on June 27, 2024.}
    \label{fig:lidar_boundaries}
\end{figure}

The 3DEP data are well-suited for this study for two reasons: 1) its base specifications are designed for consistent data acquisition and the production of derived products, allowing the entire collection to be treated as a unified ``3DEP'' dataset; and 2) it captures the U.S.’s diverse land cover and varied terrain, making it an excellent foundation for constructing robust and versatile models with broad applicability across a wide range of geospatial and environmental contexts.
The point cloud data are accessible via AWS~\cite{usgs_3dep_lidar_aws}, allowing for programmatic downloads. Additionally, each LiDAR point cloud includes boundary data, enabling users to easily define and select their area of interest. The point cloud boundaries used in this study are depicted in Figure~\ref{fig:lidar_boundaries}.

\subsection{Geospatial sampling}\label{sec: geospatial}
While 3DEP offers abundant resources for conducting pre-training, utilizing the entire dataset ($>$300TB) is practically infeasible. Therefore, a sampling strategy is required to extract representative data from U.S. regions while ensuring pre-training remains feasible. Motivated by the fact ALS is often used for analyzing terrain and land cover, we designed the sampling method to maximize diversity in both land cover and terrain.

\subsubsection{Land cover} 

We use the NLCD as the source of land cover information. The NLCD product suite provides comprehensive data on nationwide land cover and changes over two decades (2001–2021), offering detailed, long-term insights into land surface dynamics.

In this study, we utilize the latest NLCD2021 release, which includes land cover maps for 2001, 2004, 2006, 2008, 2011, 2013, 2016, 2019, and 2021. The NLCD2021 follows the same protocols and procedures as previous releases, ensuring compatibility with the 2019 database. As a result, analysis conducted for NLCD 2019 can be useful for understanding NLCD2021. For instance, the validation report for the 2019 release~\cite{wickham2023thematic} is used to understand classification accuracy of the land cover classes, which will later inform the selection of reliable land covers for data downloads.


The NLCD land cover product uses an adapted version of the Anderson Level II classification system, which includes 16 land cover classes (excluding those specific to Alaska). This system is derived from the original Anderson land use and land cover classification framework~\cite{anderson1976land}, designed to balance compatibility with U.S. federal classification systems, distinguishability of classes using primarily remote sensing data, and a hierarchical structure among classes. In this study, we use the Level I classification system by merging Level II classes, addressing the moderate per-class accuracy reported for the Level II system~\cite{wickham2023thematic}. The Level I and Level II classification systems are presented in Table~\ref{tab: classification_system}, with detailed definitions available in~\cite{nlcd_legend}.

\begin{table}[htb]
\centering
\begin{tabular}{|l|l|}
\hline
Level I Class & Level II Class \\ \hline
Water & 11: Open Water  \\ 
& 12: Perennial Ice/Snow  \\ \hline
Developed & 21: Developed, Open Space  \\ 
& 22: Developed, Low Intensity  \\ 
& 23: Developed, Medium Intensity  \\ 
& 24: Developed, High Intensity  \\ \hline
Barren & 31: Barren Land (Rock/Sand/Clay)  \\ \hline
 Forest & 41: Deciduous Forest  \\ 
& 42: Evergreen Forest  \\ 
& 43: Mixed Forest  \\ \hline
 Shrubland & 51: Dwarf Scrub (Alaska only) \\ 
& 52: Shrub/Scrub  \\ \hline
Herbaceous & 71: Grassland/Herbaceous  \\ 
& 72: Sedge/Herbaceous (Alaska only) \\ 
& 73: Lichens (Alaska only)  \\ 
& 74: Moss (Alaska only)  \\ \hline
Planted/Cultivated & 81: Pasture/Hay  \\ 
& 82: Cultivated Crops  \\ \hline
Wetlands & 90: Woody Wetlands  \\ 
& 95: Emergent Herbaceous Wetlands  \\ \hline
\end{tabular}
\caption{Level I and Level II Land Cover Classification System for NLCD2021.}
\label{tab: classification_system}
\end{table}

\subsubsection{Terrain}\label{sec: terrain_info}
We utilize the seamless DEM provided by the National Map to obtain elevation data. The seamless DEM is a high-quality geospatial dataset developed by the USGS to support a wide range of geospatial research and applications. It integrates elevation data from various sources, including airborne LiDAR, photogrammetry, and cartographic contours, into a standardized format. Available at multiple resolutions—1/3 arc-second (about 10 meters), 1 arc-second (about 30 meters), and 2 arc-seconds (about 60 meters)—these DEMs provide flexibility for detailed local analyses and broader regional studies.

\begin{table}[htb]
\centering
\begin{tabular}{lll}
\hline
Slope Class & Degree & Percentage \\ \hline
Flat        & $0^\circ - 5^\circ$      & $0\% - 8.7\%$       \\ 
Sloped     & $5^\circ - 17^\circ$     & $8.7\% - 30.6\%$    \\ 
Steep       & $\geq 17^\circ$          & $\geq 30.6\%$       \\ \hline
\end{tabular}
\caption{Slope classification with degree and percentage ranges.}
\label{tab: slope_classification_percentage}
\end{table}

For this study, we use the 1 arc-second DEM to match the resolution of the land cover maps for joint analysis. The DEM is converted into a slope classification map to reflect topographic complexity. This slope map is further classified according to the system shown in Table~\ref{tab: slope_classification_percentage}, adapted from the USDA classification~\cite{slope_classification}, to facilitate the joint analysis of land cover and terrain. The dataset is downloaded programmatically using the py3dep Python package~\cite{py3dep}.

\subsubsection{Data sampling}
Due to the highly skewed distribution of land surface elements and the sheer data volume of 3DEP LiDAR point clouds, sampling is necessary to create a diversified dataset while keeping pre-training feasible. However, random sampling risks over-representing prevalent landscapes, such as forests, while under-representing less common features, leading to an unbalanced dataset.

\begin{figure}[htb]
    \centering
    \includegraphics[width=\textwidth]{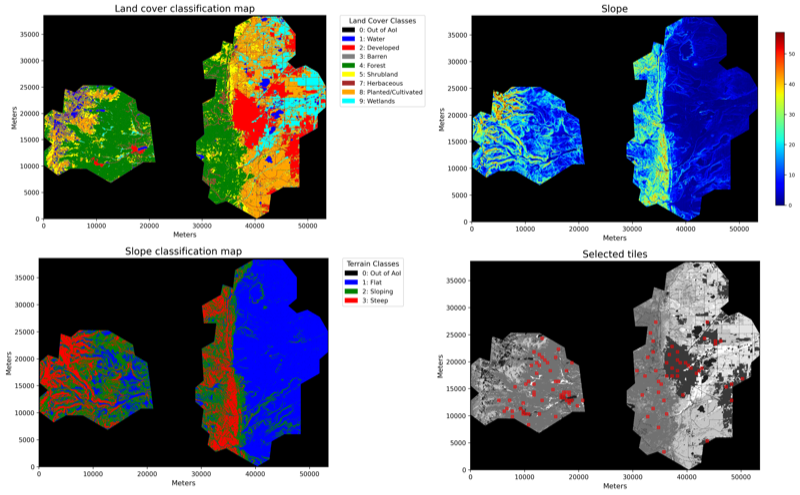} 
    \caption{The upper left figure displays the land cover map derived from the Anderson Level 1 classification system, while the upper right figure shows the slope derived from the DEM. The lower left figure presents the slope classification map, and the lower right figure illustrates the locations of the sampled tiles based on our sampling strategy.}
    \label{fig:sample_examples}
\end{figure}

To this end, we develop a geospatial sampling method to create a dataset representative of diverse land cover and terrain types. This method leverages land cover and slope classification maps to select point cloud tiles from the 3DEP LiDAR database. Specifically, we first extract a land cover map from NLCD2021 by aligning the map year with the point cloud capture year. If an exact match is unavailable, the map from the closest year is used. For each point cloud project in 3DEP, the land cover map and DEM are projected to the local Universal Transverse Mercator (UTM) coordinate reference system (CRS). The point cloud boundary file is then used to crop both maps, reducing computational complexity. 
Next, each cropped region is divided into $500\,\text{m} \times 500\,\text{m}$ patches, and the most frequent land cover and slope classes within a patch are assigned as labels. We apply inverse probability sampling on the joint probability distribution of land cover and slope classes to ensure balanced sampling across all combinations. For land cover, we focus exclusively on the "Developed" and "Forest" classes, as their combinations with slope classes effectively address common ALS downstream tasks. Additionally, these classes are among the most reliable land cover classifications, as reported in the NLCD2019 accuracy assessment~\cite{usgs_nlcd_2019_accuracy}.
After extracting patches from all point clouds, we record the bounding box coordinates for each patch, facilitating programmatic downloads of selected point cloud patches from the 3DEP database hosted on AWS. Geospatial operations such as reprojection are conducted using the rasterio and GeoPandas Python libraries, while PDAL is used for downloading data from the 3DEP server. The examples of the land cover map, slope map, slope classification map, and the sampling results are shown in Figure~\ref{fig:sample_examples}.

\subsection{Dataset statistics}\label{sec: dataset_stats}

\begin{table}[htb]
\centering
\begin{tabular}{llll}
\hline
       & Developed & Forest & All   \\ \hline
Flat   & 28,523     & 18,071  & 46,594 \\
Sloped & 3,774      & 16,021  & 19,795 \\
Steep  & 308       & 7,065   & 7,373  \\
All    & 32,605     & 41,157  & 73,762 \\ \hline
\end{tabular}
\caption{Number of tiles for each land cover and terrain class.}
\label{tab:number_of_tiles}
\end{table}


\begin{table}[htb]
\centering
\begin{tabular}{lccc}
\hline
    Dataset
    & Year
    & Coverage
    & \#Points
    \\
\hline
    ISPRS~\cite{rottensteiner2014results} 
    & 2014 
    & -- 
    & 1.2 M 
    \\
    DublinCity~\cite{zolanvari2019dublincity}
    & 2019 
    & $2 \times 10^6 \, \text{m}^2$ 
    & 260 M 
    \\
    LASDU~\cite{ye2020lasdu}
    & 2020 
    & $1.02 \times 10^6 \, \text{m}^2$
    & 3.12 M 
    \\
    DALES~\cite{dales}
    & 2020 
    & $10 \times 10^6 \, \text{m}^2$ 
    & 505 M 
    \\
    ECLAIR~\cite{melekhov2024eclair}
    & 2024 
    & $10.3 \times 10^6 \, \text{m}^2$ 
    & 582 M 
    \\
\hline
    IDTReeS~\cite{graves2020idtrees}
    & 2021 
    & $3440 \, \text{m}^2$ 
    & 0.02 M
    \\
    PureForest~\cite{pureforest}
    & 2024 
    & $339 \times 10^6 \, \text{m}^2$ 
    &  15 B
    \\
\hline
    ISPRS filtertest~\cite{sithole2004experimental} 
    & -- 
    & $1.1 \times 10^6 \, \text{m}^2$ 
    & 0.4 M
    \\
    OpenGF~\cite{qin2021opengf}
    & 2021 
    & $47.7 \times 10^6 \, \text{m}^2$ 
    & 542.1 M
    \\
\hline
    3DEP (Ours)
    & - 
    & $17691 \times 10^6\, \text{m}^2$
    & 184 B
    \\
\hline
\end{tabular}
\caption{Specifications of representative geospatial datasets. 
} 
\label{tab:geospatial_datasets}
\end{table}

Based on the sampling strategy outlined above, we constructed the dataset. Although the number of tiles per LiDAR point cloud (or a LiDAR project) is limited to 40 in our experiments, this approach can be scaled to include any number of tiles, up to covering the entire area of the LiDAR projects. The distribution of tiles across classes is summarized in Table~\ref{tab:number_of_tiles}. 
A total of 73,762 tiles are sampled, with Flat areas being the most dominant category, accounting for 63.2\%. 
In contrast, Sloped areas contribute 19,795 tiles (26.8\%), where Forest dominates with 16,021 tiles compared to only 3,774 tiles for Developed areas. 
The Steep areas, representing the smallest category, account for 7,373 tiles (10.0\%), with the majority (7,065 tiles) being Forest and only 308 tiles are classified as Developed. This distribution highlights the tendency for Developed land cover to be concentrated in Flat areas, while Forest land cover extends significantly into Sloped and Steep areas, where development is minimal. Overall, the Developed and Forest tiles are balanced. Therefore, our sampling strategy ensures a relatively balanced representation of different land cover whereas the topographic classes are limited by real-world conditions where steep slopes are predominantly forested, and development occurs primarily on flat terrain. 

Table~\ref{tab:geospatial_datasets} lists some representative ALS datasets for comparison. As shown, our dataset is the largest in terms of both geographical coverage and the number of points. Additionally, while other datasets are often specialized for specific target tasks and therefore include limited land cover or terrain types, our dataset encompasses a diverse range of land cover and terrain types at scale. Although the OpenGF dataset includes various land cover and terrain types, its geographical coverage is significantly smaller compared to ours, making it less suitable for large-scale pre-training and fine-tuning paradigms. 

Furthermore, we analyze key characteristics of the constructed dataset, including point density per square meter, ground point standard deviation, and return attributes. Due to the dataset’s large size, we conduct this analysis on a subset created through random sampling. Specifically, we randomly select 30\% of the dataset, amounting to 22,129 tiles.

\begin{table}[htb]
\centering
\begin{tabular}{llll}
\hline
       & Developed & Forest    & All       \\ \hline
Flat   & 7.7/9.9  & 11.1/15.3 & 9.0/12.4  \\
Sloped & 11.2/16.0 & 11.9/15.5 & 11.8/15.6 \\
Steep  & 28.0/39.6 & 18.0/19.2 & 18.4/20.5 \\
All    & 8.2/11.6  & 12.6/16.3 & 10.7/14.6 \\ \hline
\end{tabular}
\caption{Density per square meter for each land cover and terrain classes. The average density and their standard deviations are reported.}
\label{tab:density}
\end{table}

Table~\ref{tab:density} presents the mean and standard deviation (mean/std) of density values across land cover types (Developed and Forest) and slope categories (Flat, Sloped, and Steep). Forested areas consistently demonstrate higher mean densities and greater variability compared to Developed areas in both the Flat and Sloping categories. This is a natural outcome of multiple returns caused by vegetation layers, such as tree canopies, compared to the smoother, engineered surfaces typically found in Developed areas.
The high mean and variability observed in the Developed and Steep class may be attributed to the following factors: 1) although labeled as Developed, a majority of the tiles in this class primarily contain vegetation with minimal artificial structures; and 2) the presence of high-density LiDAR projects, which elevate both the mean and standard deviation values.
Overall, Forested areas exhibit higher densities than Developed areas across all topographies, highlighting the impact of vegetation and natural irregularities. Additionally, the observed increase in density from Flat to Steep terrain aligns with growing terrain complexity, validating the dataset’s alignment with real-world characteristics. 

\begin{table}[htb]
\centering
\begin{tabular}{llll}
\hline
       & Developed & Forest    & All       \\ \hline
Flat   & 2.5/3.4   & 4.9/4.1   & 3.5/3.8   \\
Sloped & 12.7/6.1  & 14.7/8.7  & 14.4/8.3  \\
Steep  & 36.1/18.8 & 43.6/19.1 & 43.3/19.1 \\
All    & 4.0/6.1   & 15.3/16.8 & 10.4/14.4 \\ \hline
\end{tabular}

\caption{Standard deviation of ground points for each land cover and terrain classes. The average and standard deviations are reported.}
\label{tab:ground_std}
\end{table}

Table~\ref{tab:ground_std} summarizes the mean and standard deviation (mean/std) of ground point standard deviation across land cover types and topographic categories. The results show that forested areas consistently exhibit greater variability in ground elevation compared to developed areas across all topographies. This disparity is expected, as forested areas often feature irregular terrain, whereas developed areas are typically engineered for smoothness.
Moreover, the observed increase in variability from flat to steep terrain aligns with expectations, indicating meaningful distinctions between land cover and topographic categories.

\begin{table}[htb]
\centering
\begin{tabular}{lrr}
\hline
Return number & Sum Point Count & Percent(\%) of Total \\ \hline
Single        & 13,036,774,317   & 67.92                \\
First         & 15,495,147,301   & 80.73                \\
First of many & 2,446,338,594     & 12.74                \\
Second        & 2,491,508,834   & 12.98              \\
Third         & 839,493,704    &4.37                 \\
Fourth        & 255,303,252    &1.33                 \\
Fifth         & 71,200,968    &0.37                 \\
Sixth         & 28,103,502    & 0.15                 \\
Seventh       & 14,103,705    & 0.07                 \\
Last          & 15,594,824,929   & 81.24                \\
Last of many  & 2,559,449,009   & 13.33                \\ \hline
\end{tabular}
\caption{Return characteristics of the Developed class}
\label{tab:return_dev}
\end{table}

\begin{table}[htb]
\centering
\begin{tabular}{lrr}
\hline
Return number     & Sum Point Count  & Percent(\%) of Total \\ \hline
Single            & 27,509,188,124      & 49.32                \\ 
First             & 38,520,832,462      & 69.07                \\
First of Many     & 10,992,636,342   & 19.71                \\
Second            & 11,082,071,467   & 19.87                \\
Third             & 4,272,395,407    &7.66                 \\
Fourth            & 1,369,446,735    &2.46                 \\
Fifth             & 364,742,990    &0.65                 \\
Sixth             & 115,577,887    &0.21                 \\
Seventh           & 47,218,274    &0.08                 \\
Last              & 38,817,185,224   &69.60                \\
Last of Many      & 11,310,467,287   &20.28                \\
\hline
\end{tabular}
\caption{Return characteristics of the Forest class}
\label{tab:return_forest}
\end{table}

Table~\ref{tab:return_dev} summarizes the return characteristics of the Developed class, highlighting the predominance of ``Last'' and ``First'' returns, with ``Single'' returns also representing a significant portion. Intermediate returns, such as “Second” and “Third,” along with higher-order returns, contribute only minimally. The “First of Many” and “Last of Many” categories account for 12.74\% and 13.33\%, respectively, showing the relatively minor role of multi-return sequences. This distribution reflects the predominance of simpler, standalone geometries or terminal interactions within the Developed class point clouds.
Table~\ref{tab:return_forest} presents the return characteristics of the Forest class. “Last” and “First” returns dominate, each representing approximately 69\%, while “Single” returns account for 49.32\%. Intermediate returns, such as “First of Many” and “Second,” suggest significant interactions with the canopy, whereas higher-order returns collectively contribute minimally. Similar to the Developed class, this distribution reflects the dominance of single and terminal returns from the canopy and ground, with limited deeper returns, aligning with the vertical structure characteristic of forested environments.

The comparison of return characteristics between Developed and Forest classes highlights notable differences. While “Last” and “First” returns dominate in both, their proportions are significantly higher in Developed compared to Forest, alongside a larger share of “Single” returns. These differences reflect the smoother, reflective surfaces in developed areas, which produce simpler returns. Conversely, the Forest class shows a greater proportion of intermediate returns, indicating more complex canopy interactions. 

The dataset’s density, ground variability, and return characteristics exhibit expected trends across various land cover types and topographic categories, thereby confirming the validity of the sampling method. These results collectively demonstrate that the sampling approach effectively captures the essential spatial and structural features of the LiDAR data, ensuring the constructed dataset accurately reflects real-world conditions.

\begin{figure}[htb]
    \centering
    \includegraphics[width=\textwidth]{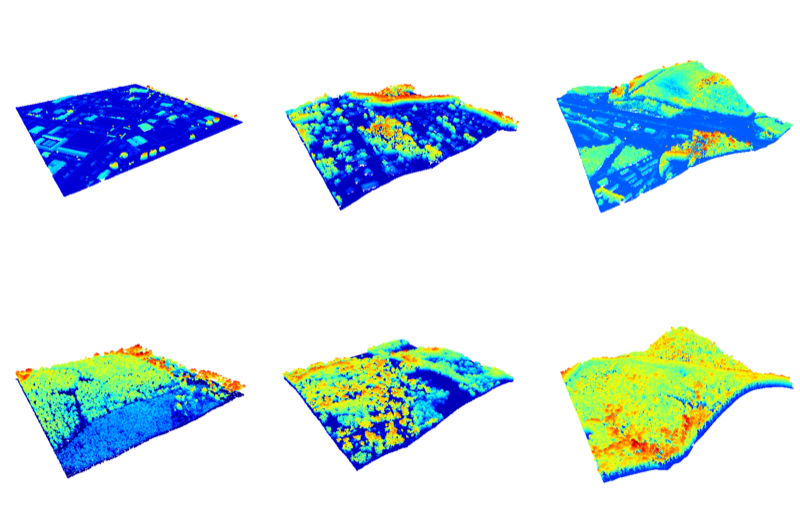} 
    \caption{Random samples of the dataset. Top: point cloud tiles labeled as ``Developed''. Bottom: point cloud tiles labeled as ``Forest''. From left to right: point cloud tiles labeled as ``Flat'', ``Sloping'', ``Steep''.}
    \label{fig:sample_point_clouds}
\end{figure}

Additionally, random samples of the extracted point cloud tiles are presented in Figure~\ref{fig:sample_point_clouds}. Developed tiles include built-up areas such as cities and villages, while Forest tiles primarily consist of vegetation. Furthermore, the increasing slope classes illustrate the growing complexity of terrain conditions, transitioning from flat to steep regions.

\section{Model architecture and pre-training}\label{sec: model}
We adopt BEV-MAE~\cite{bevmae} as our pre-training method. BEV-MAE is a cutting-edge pre-training method for 3D point clouds. The method is originally proposed for autonomous driving. It is designed for outdoor data and is designed to handle  data from a Birds-Eye-View (BEV) perspective. The BEV-MAE inherits its design from Masked AutoEncoders (MAE)~\cite{mae} in 2D image processing, which is an SSL technique that learns the representation by masking out a large portion of data and reconstructing them. This way of pre-training is assumed to be helpful for the model to learn high-level representation by forcing the model to reconstruct the invisible parts. 

BEV-MAE follows a similar way of processing pipeline as MAE: masking, backbone network, and reconstruction. An element of the mask (corresponding to a masked patch in image processing) in BEV-MAE is a 3D pillar (or a BEV cell) that encapsulates a volume of 3D points. Specifically, given a defined x and y ranges, all points that fall in the range are removed. All masked elements are replaced by the same learnable mask token. After masking, the remaining point clouds together with mask tokens are put into the encoder network. the encoder network is a 3D sparse CNN~\cite{subsparse, minknet}, which is an memory efficient variant of voxel-based 3D CNN. It learns the multi-scale representation of point clouds with successive convolutions and downsampling. After the encoder, the masked point clouds are reconstructed from the corresponding mask tokens with a series of convolutions. In addition, the average number of points for each masked region is also reconstructed to learn about the local density. 

We consider BEV-MAE to be suitable for this study for the following reasons. Unlike other MAEs for point clouds such as MaskPoint~\cite{liu2022masked} and Point-MAE~\cite{pang2022masked}, BEV-MAE is explicitly designed for handling outdoor data. Second, BEV-MAE adopts sparse CNN as its backbone which is more scalable to large-scale point clouds compared to Point-based methods. Last but not least, it is explicitly designed to utilize BEV perspective which is even more natural for ALS as it fits the way ALS point clouds are captured. 

For pre-training, we use the AdamW optimizer with $\beta_1 = 0.9$, $\beta_2 = 0.99$, a batch size of 16, and a one-cycle cosine annealing scheduler with a maximum learning rate of $10^{-2}$. The models are trained for 50 epochs, with each epoch exposing the entire training dataset to the model once. We increase the number of parameters of the model to 60 M by increasing the channel widths without changing the depths. 

The input point cloud is a square tile with a side length of 500 meters. During training, we randomly crop smaller $144\,\text{m} \times 144\,\text{m}$ tiles from the original tile. Each cropped tile is voxelized with a voxel size of $0.6\,\text{m}$, and up to 200,000 voxels are sampled, with each voxel containing a maximum of 5 points. For the ground truth of point coordinate reconstruction, the point cloud is voxelized using a voxel size of $4.8\,\text{m} \times 4.8\,\text{m} \times 288\,\text{m}$ to create BEV voxels, where the maximum number of voxels is set to 200,000 and each voxel can contain up to 30 points. The ground truth density is computed on the fly from the BEV voxels. Several common data augmentations including random flipping, scaling, and translation are performed for the input point clouds.  

The input features consist of the 3D coordinates of the points. All pre-training runs were conducted on the AI Bridging Cloud Infrastructure (ABCI) 2.0 using up to 4 NVIDIA V100 GPUs.

\begin{figure}[htp]
    \centering
    \includegraphics[width=\textwidth]{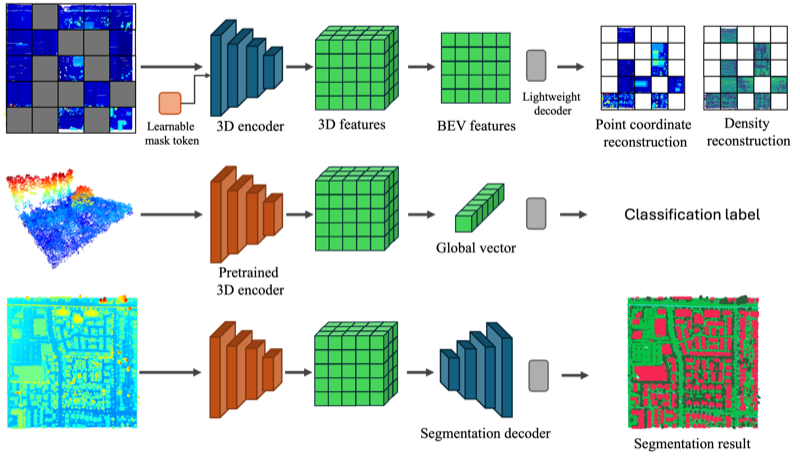} 
    \caption{Overview of the pre-training and fine-tuning using BEV-MAE.}
    \label{fig:bevmae_overview}
\end{figure}

\section{Experiments}\label{sec: experiments}
We evaluate the pre-trained model on several downstream tasks, specifically focusing on tree species classification, terrain scene recognition, and point cloud semantic segmentation. In the following sections, we introduce the tasks, datasets, and fine-tuning architectures used for these downstream tasks. We then describe the evaluation metrics used in this study.

\subsection{Task}
\subsubsection{Tree species classification}
Tree species classification is a vital task for managing forests. The identification of tree species supports public policies for forest management and helps mitigate the impact of climate change on forests. To validate the effectiveness of the approach for tree species classification, we use the PureForest~\cite{pureforest} dataset, a comprehensive collection tailored for analyzing forest environments. The dataset comprises 135,569 patches, each measuring $50\,\text{m} \times 50\,\text{m}$, and covers a total area of $339\,\text{km}^2$ across 449 distinct closed forests located in 40 departments in southern France. It includes 18 tree species, categorized into 13 semantic classes: Deciduous oak, Evergreen oak, Beech, Chestnut, Black locust, Maritime pine, Scotch pine, Black pine, Aleppo pine, Fir, Spruce, Larch, and Douglas. The dataset provides two modalities: colored ALS point clouds (40 points/m²) and aerial images (spatial resolution of $0.2\,\text{m}$). The task is to classify a patch into one of the 13 semantic categories. We use only the 3D point coordinates as input. Since each tile measures $50\,\text{m} \times 50\,\text{m}$, we use the entire tile as input during both training and testing. Following the pre-training settings, we limit the number of voxels after voxelization to 200,000. All the other settings remain the same as the pre-training ones.

The architecture used for this task is illustrated in Figure~\ref{fig:bevmae_overview} (middle). The model outputs a classification label representing the tree species of the input patch. To achieve this, the encoder's output is spatially pooled to form a global vector that summarizes the input point cloud. We concatenate the average-pooled and max-pooled vectors to emphasize both sharp and smooth features. The resulting global descriptor is then classified through a series of fully connected layers.

\subsubsection{Terrain scene recognition}
3D terrain scene recognition is crucial for understanding and classifying landforms, which plays a significant role in geography-related research areas such as digital terrain analysis and ecological environment studies~\cite{qin2018deep}. Therefore, it is vital to validate the effectiveness of the pre-trained model on terrain scene classification. However, there are very few publicly available datasets for this task. While a prior study~\cite{qin2018deep} exists, the data used in the study remains private. 

To address this limitation, we develop our own dataset to evaluate our model on terrain scene recognition. We base our dataset on OpenGF~\cite{qin2021opengf}, which was originally designed for ground filtering. In OpenGF, the authors divided the data into four prime terrain types—Metropolis, Small City, Village, and Mountain—consisting of 160 $500\,\text{m} \times 500\,\text{m}$ point cloud tiles for training and validation. These terrain types are further subdivided into nine scenes: Metropolis is divided into regions with large roofs (S1) and dense roofs (S2); Small City is divided into tiles with flat ground (S3), locally undulating ground (S4), and rugged ground (S5); Village consists only of tiles with scattered buildings (S6); Mountain is divided into tiles with gentle slopes and dense vegetation (S7), steep slopes and sparse vegetation (S8), and steep slopes and dense vegetation (S9). Given the well-defined scene categories and the large size of the dataset, we create a dataset for terrain scene recognition based on OpenGF.

\begin{figure}[htp]
    \centering
    \includegraphics[width=0.8\textwidth]{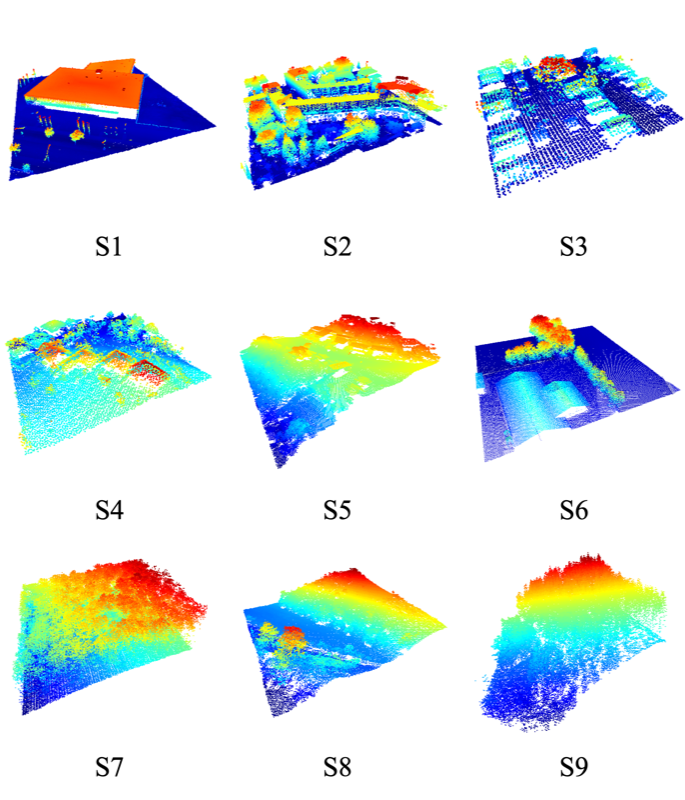} 
    \caption{Some examples from the terrain scene recognition dataset developed based on OpenGF. S1 and S2 refer to metropolitan areas with large and dense roofs, respectively. S3 corresponds to a small city with flat ground, while S4 and S5 represent small cities with locally undulating ground and rugged ground, respectively. S6 denotes village areas, S7 corresponds to mountain areas with gentle slopes and dense vegetation, S8 represents mountain areas with steep slopes and sparse vegetation, and S9 indicates steep slopes with dense vegetation. The points are color-coded by elevation.} 
    \label{fig:terrain_scene_recognition_example}
\end{figure}

To construct the dataset, we first combine the training and validation tiles from OpenGF. We then split the combined tiles into training, validation, and test sets, assigning 106 tiles (about 66\%) to training, 27 tiles (about 17\%) to validation, and 27 tiles (about 17\%) to testing. For the training tiles, we divide each $500\,\text{m} \times 500\,\text{m}$ tile into smaller $100\,\text{m} \times 100\,\text{m}$ tiles using a sliding window cropping algorithm with an overlap of $50\,\text{m}$. For validation and test tiles, no overlap is applied. This process results in 10,591 tiles for training, 675 tiles for validation, and 675 tiles for testing. Examples of the dataset are shown in Figure~\ref{fig:terrain_scene_recognition_example}.

The task is to classify a patch into one of the nine semantic categories (S1--S9). Similar to the tree species classification task, we use only 3D coordinates as input. During training, the entire tile is fed into the network. Following the pre-training settings, we limit the number of voxels after voxelization to 200,000. All the other settings remain the same as the pre-training ones. For the fine-tuning architecture, we use the same architecture as that used for tree species classification.

\subsubsection{Point cloud semantic segmentation}
Urban point cloud semantic segmentation provides vital information about ground objects for urban modeling. It involves classifying points in 3D data into meaningful categories such as buildings, roads, vegetation, and other urban features. In this work, we use the Dayton Annotated Laser Earth Scan (DALES) dataset~\cite{dales}, an aerial LiDAR dataset with nearly half a billion points spanning 10 square kilometers, to evaluate the performance of the pre-trained models. DALES consists of 40 scenes of dense, labeled aerial data covering multiple scene types, including urban, suburban, rural, and commercial. The dataset is hand-labeled by expert LiDAR technicians into eight semantic categories: ground, vegetation, cars, trucks, poles, power lines, fences, and buildings. While sensor intensity and return information are available, we use only the $x$, $y$, and $z$ features as input. Each patch measures $500\,\text{m} \times 500\,\text{m}$, and the task is to classify points into one of the semantic classes. Similar to the pre-training setup, we sample $144\,\text{m} \times 144\,\text{m}$ tiles from the $500\,\text{m} \times 500\,\text{m}$ patches during training. The maximum number of voxels is set to 200,000 during training and 1,000,000 during testing to ensure that all points are classified. All the other settings remain the same as the pre-training ones.

The architecture used for this task is shown in Figure~\ref{fig:bevmae_overview} (bottom). The model outputs a point cloud with per-point classification labels. To achieve this, we append a decoder to the pre-trained BEV-MAE encoder to recover the full resolution of the points. The decoder receives the downsampled high-level representations from the encoder and gradually transforms and upsamples the point cloud until it reaches full resolution. The resulting architecture is a U-Net~\cite{unet}-style 3D CNN, which connects the encoder and decoder using skip connections. After passing through the decoder, each point is classified using a series of fully connected layers.

\subsection{Evaluation Metrics}\label{sec:eval_metrics}

We use two commonly adopted metrics to evaluate the performance of the models: Mean Intersection over Union (mIoU) and Overall Accuracy (OA). 

mIoU evaluates a model’s performance by measuring the overlap between predicted and ground truth points or point clouds. The IoU for each class $i$ is defined as:
\begin{equation}
\text{IoU}_i = \frac{\text{TP}_i}{\text{TP}_i + \text{FP}_i + \text{FN}_i},
\end{equation}
where $C$ is the total number of classes, $\text{TP}_i$ represents the true positives for class $i$, and $\text{FP}_i$ and $\text{FN}_i$ denote the false positives and false negatives, respectively. IoU is considered a stricter and more comprehensive metric compared to metrics like precision and recall, as it penalizes both over-prediction ($\text{FP}_i$) and under-prediction ($\text{FN}_i$). The mIoU is then calculated as the average IoU across all classes:
\begin{equation}
\text{mIoU} = \frac{1}{C} \sum_{i=1}^{C} \text{IoU}_i.
\end{equation}
mIoU averages performance across classes, ensuring that both major and minor classes are equally weighted. Consequently, a high mIoU requires the model to perform well on all classes, regardless of their prevalence in the dataset.

OA measures the proportion of correctly classified points or point clouds across all classes. Mathematically, OA is defined as:
\begin{equation}
\text{OA} = \frac{\sum_{i=1}^{C} \text{TP}_i}{\# \text{all points or point clouds}}.
\end{equation}
OA does not account for class imbalance, meaning that a high OA can be achieved by performing well on major classes, even if performance on minor classes is poor.

These two metrics complement each other: mIoU emphasizes the model’s ability to accurately predict each class, providing a balanced evaluation of model performance, while OA captures the overall classification accuracy across all classes.

\section{Results}\label{sec: results}

In this section, we present the pre-training results through the reconstruction performance of BEV-MAE. We then discuss the results of various downstream tasks, including tree species classification, terrain scene recognition, and point cloud semantic segmentation.

\subsection{Pre-training results}\label{sec:pre-training result}
\begin{figure}[htb]
    \centering
    \includegraphics[width=1.0\textwidth]{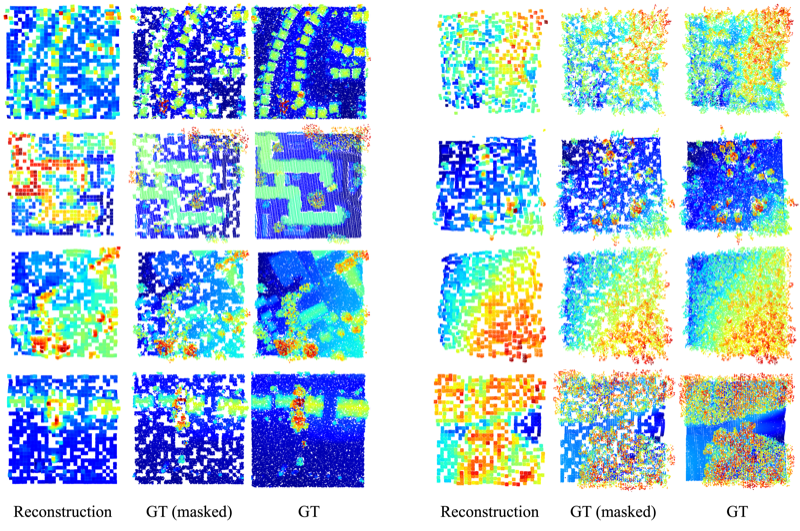} 
    \caption{Results of coordinate reconstruction. GT represents the Ground Truth, while GT (masked) displays only the ground truth point clouds within the masked regions. }
    \label{fig:coord_recon}
\end{figure}

To validate the quality of the learned representations, we visualize the reconstructed coordinates and densities in this section. We feed unseen point cloud tiles (not used during pre-training) into the model and visualize the reconstructed coordinates and densities. 

As shown in Figure~\ref{fig:coord_recon}, our first observation is that the model effectively reconstructs the overall patterns of the point clouds. The reconstructed surface objects, including buildings, trees, and the ground, align roughly well with the ground truth point clouds. However, a significant amount of detail is missing in the reconstructions, suggesting that detailed shape information is not fully recovered. Specifically, points within individual BEV cells often reconstruct as simple plane-like shapes, failing to capture the fine-grained details of the objects. This indicates that the network primarily learns abstract shapes of the point clouds rather than their fine-grained geometry. Furthermore, this limitation suggests the model may struggle to recognize smaller objects, as their relatively small size and detailed shape information are more easily obscured during the masked autoencoding process. 

\begin{figure}[htb]
    \centering
    \includegraphics[width=1.\textwidth]{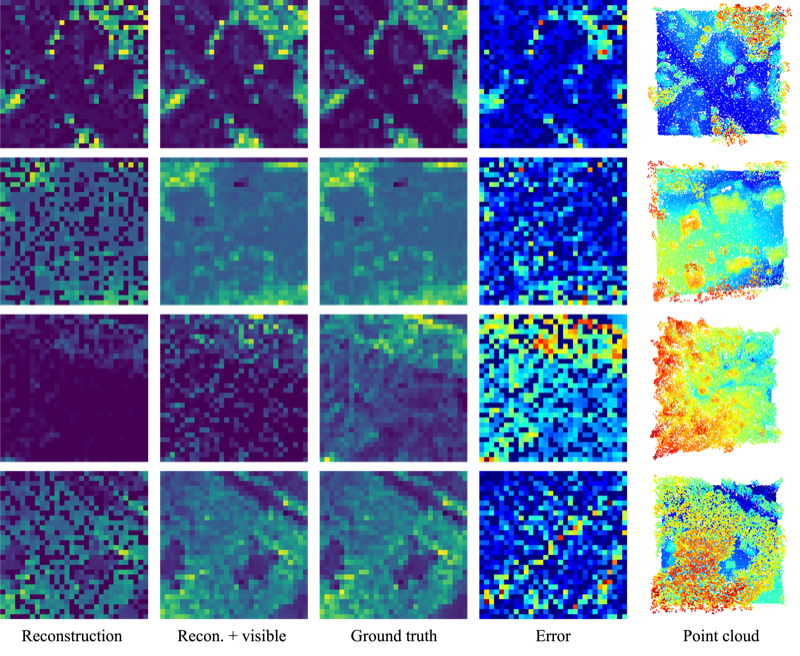} 
    \caption{Results of density reconstruction.}
    \label{fig:recon_density}
\end{figure}

Figure~\ref{fig:recon_density} shows the results of density reconstruction. In general, the model achieves high-quality reconstruction. The “Recon. + visible” outputs often closely mimic the ground truth. The error maps reveal that errors are frequently concentrated in regions where density is higher. For instance, in the first row of the figure, errors are primarily concentrated around trees, whereas errors at ground or building points are minimal. Similarly, the third row shows that regions with highly variable topography under the dense vegetations (upper right) pose challenges for density reconstruction.

\subsection{Downstream application results}
\subsubsection{Tree species classification}
\begin{table}[htb]
\centering
\begin{tabular}{lcc}
\hline
Method & mIoU (\%) & OA (\%) \\
\hline
Lidar (Baseline)~\cite{pureforest} & 55.1 & 80.3 \\
Lidar + RGBI~\cite{pureforest} & 53.6 & 79.1 \\
Lidar + Elevation~\cite{pureforest} & 57.2 & 83.6 \\
Aerial Imagery~\cite{pureforest} & 50.0 & 73.1 \\
\hline
BEV-MAE (Scratch) & 72.2 & 86.8 \\
BEV-MAE (Ours) & \textbf{75.6} &  \textbf{87.1} \\
\hline
\end{tabular}
\caption{Results of tree species classification. Our scores are the average of three runs. Bold text shows the best performance.}
\label{tab:tree_species_comparison_overall}
\end{table}

\begin{table}[htb]
\centering
    \begin{tabular}{lcccc}
    \hline
    Category & \#Patches & Lidar (baseline) & Scratch & Ours \\
    \hline
    Deciduous oak & 48055 & 73.4 & 78.3 & \textbf{78.5} \\
    Evergreen oak & 22361 & 59.4 & 63.2 & \textbf{63.7} \\
    Beech & 12670 & 88.8 & \textbf{93.1} & 92.2 \\
    Chestnut & 3684 & 56.5 & \textbf{65.6} & 62.0 \\
    Black locust & 2303 & 58.1 & 73.7 & \textbf{82.2} \\
    Maritime pine & 7568 & 62.9 & 96.2 & \textbf{97.5} \\
    Scotch pine & 18265 & 58.6 & 86.7 & \textbf{88.2} \\
    Black pine & 7226 & 46.2 & 74.3 & \textbf{79.0} \\
    Aleppo pine & 4699 & 39.3 & 87.9 & \textbf{93.7} \\
    Fir & 840 & 0.0 & 0.0 & 0.0 \\
    Spruce & 4074 & 85.8 & \textbf{94.2} & 93.3 \\
    Larch & 3294 & 50.6 & 81.7 & \textbf{85.6} \\
    Douglas & 530 & 36.5 & 78.8 & \textbf{93.3} \\
    \hline
    Mean & -- & 55.1 & 74.9 & \textbf{77.6} \\
    \hline
    \end{tabular}
\caption{Detailed results of tree species classification. We show the best run among three runs for Ours and Scratch. Bold test shows the best performance.}
\label{tab:tree_species_comparison_with_patches}
\end{table}

As shown in Table~\ref{tab:tree_species_comparison_overall}, the pre-trained model outperforms the scratch model by 3.4\% in mIoU and 0.3\% in OA, highlighting the effectiveness of the dataset and pre-training for tree species classification. Specifically, in Table~\ref{tab:tree_species_comparison_with_patches}, the pre-trained model demonstrates either slight or substantial improvements over the scratch model across nearly all categories. This suggests that pre-training enables the model to learn generalizable shape-related features beneficial for distinguishing between tree species.

The most notable improvements are observed in the Black locust and Douglas classes, with performance gains of 8.5\% and 14.5\%, respectively. We hypothesize that these significant improvements stem from these species being native to the United States and likely included in the pre-training dataset. Furthermore, the improvement in the Douglas class may also be attributed to its small sample size (as shown in the Patches column), which limits the scratch model’s ability to learn transferable features effectively.        

Moreover, our models including both scratch and pre-trained ones have largely outperformed the baseline models reported by ~\cite{pureforest} even when the baseline has used additional features such as colors. Apart from the pre-training, such a big difference can be attributed partially to the difference in input data handling. For instance, while the baselines subsampled point cloud with a voxel size of 0.25m, we used 0.06 m which results in much higher resolution and maintains much finer details. Therefore, we expect our model to learn much better geometric features.

\subsubsection{Terrain scene recognition}
\begin{table}[htpb]
\centering
    \begin{tabular}{lcc}
    \hline
    \textbf{Method} & \textbf{mIoU (\%)} & \textbf{OA (\%)} \\
    \hline
    BEV-MAE (Scratch) & 86.6 & 92.6 \\
    BEV-MAE (Ours) & \textbf{87.4} &  \textbf{93.1} \\
    \hline
    \end{tabular}
\caption{Results of terrain scene recognition. The scores are the average of three runs. Bold text shows the best performance.}
\label{tab:terrain_scene_recognition_overall}
\end{table}

\begin{table}[htb]
\centering
\resizebox{\columnwidth}{!}{%
\begin{tabular}{l|ll|lll|l|lll|l}
\hline
 & \multicolumn{2}{l|}{Metropolis} & \multicolumn{3}{l|}{Small city} & Village & \multicolumn{3}{l|}{Mountain} &  \\
                & S1   & S2   & S3   & S4   & S5    & S6   & S7    & S8   & S9   & mean \\ \hline
BEV-MAE (Scratch) & 75.0 & \textbf{81.7} & \textbf{97.3} & \textbf{93.4} & \textbf{100.0} & 89.3 & 98.7  & 77.3 & 80.6 & 88.2 \\
BEV-MAE (Ours)  & \textbf{77.8} & 81.2 & \textbf{97.3} & 90.8 & \textbf{100.0} & \textbf{90.4} & \textbf{100.0} & \textbf{81.6} & \textbf{85.2} & \textbf{89.4} \\ \hline
\end{tabular}%
}
\caption{Detailed results of terrain scene recognition. The scores represent the IoUs from the best model among three runs. }
\label{tab:terrain_scene_recognition_detailed}
\end{table}

The overall results of terrain scene recognition are presented in Table~\ref{tab:terrain_scene_recognition_overall}. As shown, BEV-MAE pre-trained on the 3DEP dataset significantly outperforms the scratch model in terms of both mIoU and OA. This suggests that pre-training offers valuable improvements in the quality of representation and generally enhances terrain scene recognition. The detailed classification results for each class are presented in Table~\ref{tab:terrain_scene_recognition_detailed}. In general, the pre-trained BEV-MAE performs better on terrain scenes such as Metropolis, Village, and Mountain, while showing lower performance on Small City terrain scenes. For Metropolis classes, the pre-trained model demonstrates significantly better recognition for the S1 class compared to the scratch model, indicating a stronger semantic understanding of urban scenes, particularly buildings with large roofs. Moreover, the pre-trained model achieves substantially better performance across all subclasses of the Mountain terrain scene, highlighting its ability to understand varying terrains with vegetation. While S8 and S9 share similar steep terrain characteristics, the pre-trained model effectively differentiates between sparse and dense vegetation, performing significantly better than the scratch model. This result highlights the pre-trained model's capability to capture fine-grained semantic differences across different forest scenes where complex terrain and surface object interactions occur.

\subsubsection{Point cloud semantic segmentation for developed areas}

\begin{table}[htb]
\centering
\begin{tabular}{lcc}
\hline
\textbf{Method} & \textbf{mIoU (\%)} & \textbf{OA (\%)} \\
\hline
PointNet++~\cite{qi2017pointnet++} & 68.3 & 95.7 \\
KPConv~\cite{thomas2019kpconv} & \textbf{81.1} & \textbf{97.8} \\
RandLA~\cite{hu2020randla} & 79.3 & 97.1 \\
EyeNet~\cite{yoo2023human} & 79.6 & 97.2 \\
\hline
BEV-MAE (Scratch) & 77.9 &  97.3 \\
BEV-MAE (OpenGF~\cite{qin2021opengf}) & 77.3 & 97.2 \\
BEV-MAE (Random, 10 samples/project)   & 77.6 &97.3\\
BEV-MAE (Random, 20 samples/project) & 77.8 &97.2\\
BEV-MAE (Random, 40 samples/project) & 77.7 &97.3\\

BEV-MAE (Ours, 10 samples/project) & 77.7 & 97.3\\
BEV-MAE (Ours, 20 samples/project) & 78.0 & 97.3\\
BEV-MAE (Ours, 40 samples/project) & 78.2 & 97.3 \\
\hline
\end{tabular}
\caption{Results of point cloud semantic segmentation. Bold values indicate the best performance. Random denotes the 3DEP dataset constructed with random sampling instead of the proposed geospatial sampling.}
\label{tab:urban_seg}
\end{table}

As shown in Table~\ref{tab:urban_seg}, pre-training results in only a slight increase in average mIoU, suggesting that it has a limited impact on learning useful features for the segmentation of developed areas. We hypothesize that, as discussed in Section~\ref{sec:pre-training result}, the model struggles to capture fine-grained geometric details of objects in developed areas, which are crucial for tasks like semantic segmentation. This limitation likely contributes to the modest improvement observed.

To further investigate the effectiveness of pre-training, we trained the model on datasets of varying scales and compared it against several dataset variants. First, we create dataset variants where the proposed geospatial sampling method is replaced with random sampling, meaning sampling is performed without selectively considering land covers or topographies. Second, we pre-trained the model using OpenGF~\cite{qin2021opengf}, a large-scale dataset designed for ground filtering. OpenGF is included for comparison due to its diverse land covers and topographies, which are similar in design (but not scale) to our 3DEP dataset.

Interestingly, when the number of samples per project is relatively small (10 samples/project), the pre-trained model performed slightly worse than the scratch model. We hypothesize that insufficient data further hampers the model’s ability to learn local details. However, as the dataset scale increased with our geospatial sampling method, the model's performance steadily improved, surpassing the scratch model after reaching 20 samples per project. This demonstrates that transferable features for urban scenes can be effectively learned with an increasing number of samples.

Conversely, the model pre-trained on OpenGF achieved the lowest performance, despite its inclusion of areas such as Metropolis, Small City, and Village. These results suggest that transferable features for urban scenes cannot be effectively learned when the dataset size is limited, highlighting the critical role of dataset scale in successful pre-training. 

On the other hand, the randomly sampled datasets failed to provide meaningful improvements, even as the dataset size increased. Their performance consistently remained below that of the scratch model. This reveals that both dataset scale and the sampling strategy are crucial for effective pre-training, validating the importance of a well-designed geospatial sampling method.

\section{Conclusions}\label{sec: conclusion}
In this study, we explore the use of large-scale pre-training for ALS applications. To address the lack of large-scale datasets with sufficient land cover and terrain diversity, we develope a large-scale dataset using a proposed geospatial sampling method, guided by land cover maps and DEMs. Through extensive experiments on three downstream tasks, we demonstrate that the pre-trained model consistently outperforms the scratch model. Moreover, we validated that our sampling strategy enables the method to achieve consistent performance improvements as the dataset scale increases, in contrast to random sampling.

While we observe significant improvement in tree species classification and terrain scene recognition tasks, the performance gains in point cloud segmentation remain limited. We hypothesize that this could be due to the model’s inability to capture fine-grained details, as discussed in Section~\ref{sec:pre-training result}, which hinders its ability to object boundaries. Additionally, we believe that another contributing factor to the problem is that the baseline model, originating from the general computer vision domain, may not be fully adapted to address specific challenges in ALS data, such as the drastic variations in object scales within a single scene.

To address the aforementioned issues, one possible direction for improvement is to focus on generating more detailed reconstructions. This could be achieved by extending the loss function, such as incorporating perceptual loss~\cite{tukra2023improving}, or by adopting multi-scale SSL architectures~\cite{reed2023scale}.

Another promising direction is to develop tailored SSL methods specifically designed for ALS data, capable of handling both large-scale and small-scale objects simultaneously. For example, this could involve inventing new masking strategies~\cite{ringmo} that better address the unique challenges of ALS data.

We believe that the dataset we developed, along with the methodology used for its construction, can serve as a valuable baseline for future research on large-scale ALS dataset development. Additionally, our findings provide meaningful guidance for advancing research on applying the pre-training and fine-tuning paradigm to ALS applications. Furthermore, the pre-trained model has the potential to be fine-tuned for various downstream applications or used to transfer knowledge through methods such as knowledge distillation~\cite{hinton2015distilling}. 


\bibliographystyle{elsarticle-num-names}
\bibliography{main}

\end{document}